\documentclass[11pt,a4paper]{article}
\usepackage[hyperref]{emnlp2017}
\usepackage{times}
\usepackage{latexsym}

\emnlpfinalcopy 

\usepackage{hyperref}       
\usepackage{url}            
\usepackage{booktabs}       
\usepackage{amsfonts}       
\usepackage{nicefrac}       
\usepackage{microtype}      
\usepackage{helvet}
\usepackage{amsmath}
\usepackage{graphicx}
\usepackage{fancyhdr}
\usepackage{adjustbox}
\usepackage{multirow}
\usepackage{float}
\usepackage{textcomp}
\usepackage{algorithm,algorithmic}
\usepackage{amssymb,amsthm}
\usepackage{bbm}

\title{Flexibly-Structured Model for Task-Oriented Dialogues}

\author{
Lei Shu$^1$ \thanks{Work mostly performed as an intern at Uber AI Labs}
, Piero Molino$^2$
, Mahdi Namazifar$^2$
, Hu Xu$^1$
, Bing Liu$^1$
, Huaixiu Zheng$^2$
, Gokhan Tur$^2$\\
$^1$Department of Computer Science, University of Illinois at Chicago\\
$^2$Uber AI, \\
$^1$\{lshu3, hxu48, liub\}@uic.edu, \\
$^2$\{piero, mahdin, huaixiu.zheng, gokhan\}@uber.com 
}
\date{}
\begin{document}
\maketitle
\begin{abstract}

This paper proposes a novel end-to-end architecture for task-oriented dialogue systems.
It is based on a simple and practical yet very effective sequence-to-sequence approach, where language understanding and state tracking tasks are modeled jointly with a structured copy-augmented sequential decoder and a multi-label decoder for each slot.
The policy engine and language generation tasks are modeled jointly following that.
The copy-augmented sequential decoder deals with new or unknown values in the conversation, while the multi-label decoder combined with the sequential decoder ensures the explicit assignment of values to slots.
On the generation part, slot binary classifiers are used to improve performance.
This architecture is scalable to real-world scenarios and is shown through an empirical evaluation to achieve state-of-the-art performance on both the Cambridge Restaurant dataset and the Stanford in-car assistant dataset\footnote{The code is available at \url{https://github.com/uber-research/FSDM}}\footnote{To appear in SIGDIAL 2019}.

\end{abstract}

\section{Introduction}
A traditional task-oriented dialogue system is often composed of a few modules, such as natural language understanding, dialogue state tracking, knowledge base (KB) query, dialogue policy engine and response generation. Language understanding aims to convert the input to some predefined semantic frame. State tracking is a critical component that models explicitly the input semantic frame and the dialogue history for producing KB queries. The semantic frame and the corresponding belief state are defined in terms of informable slots values and requestable slots. Informable slot values capture information provided by the user so far, e.g., \{\textit{price=cheap, food=italian}\} indicating the user wants a cheap Italian restaurant at this stage. 
Requestable slots capture the information requested by the user, e.g., \{\textit{address, phone}\} means the user wants to know the address and phone number of a restaurant.
Dialogue policy model decides on the system action which is then realized by a language generation component.

To mitigate the problems with such a classic modularized dialogue system, such as the error propagation between modules, the cascade effect that the updates of the modules have and the expensiveness of annotation, end-to-end training of dialogue systems was recently proposed~\cite[among others]{BingNAACL18,Jason17,Lowe18,msr_challenge,BingGoogle17,Pawel18,bordes2016learning,wen2016network,serban2016building}.
These systems train one whole model to read the current user's utterance, the past state (that may contain all previous interactions) and generate the current state and response.

There are two main approaches for modeling the belief state in end-to-end task-oriented dialogue systems in the literature: the \textit{fully structured} approach based on classification~\cite{wen2016network, wen2017latent}, and the \textit{free-form} approach based on text generation~\cite{leisequicity}.
The fully structured approaches~\cite{ramadan2018large, ren2018towards} use the full structure of the KB, both its schema and the values available in it, and assumes that the sets of informable slot values and requestable slots are fixed.
In real-world scenarios, this assumption is too restrictive as the content of the KB may change and users' utterances may contain information outside the pre-defined sets.
An ideal end-to-end architecture for state tracking should be able to 
identify the values of the informable slots and the requestable slots, 
easily adapt to new domains, to the changes in the content of the KB, and to the occurrence of words in users' utterances that are not present in the KB at training time, while at the same time providing the right amount of inductive bias to allow generalization.

Recently, a free-form approach called TSCP (Two Stage Copy Net)~\cite{leisequicity} was proposed.
This approach
does not integrate any information about the KB in the model architecture.
It has the advantage of being readily adaptable to new domains and changes in the content of the KB as well as solving the out-of-vocabulary word problem by generating or copying the relevant piece of text from the user's utterances in its response generation.
However, TSCP can produce invalid states (see Section \ref{sec:experiment}).
Furthermore, by putting all slots together into a sequence, it introduces an unwanted (artificial) order between different slots since they are encoded and decoded sequentially.
It could be even worse if two slots have overlapping values, like departure and arrival airport in a travel booking system.
As such, the unnecessary order of the slots makes getting rid of the invalid states a great challenge for the sequential decoder.
As a summary, both approaches to state tracking have their weaknesses when applied to real-world applications.



This paper proposes the \underline{F}lexibly-\underline{S}tructured \underline{D}ialogue \underline{M}odel~(FSDM) as a new end-to-end task-oriented dialogue system. 
The state tracking component of FSDM has the advantages of both fully structured and free-form approaches while addressing their shortcomings.
On one hand, it is still structured, as it incorporates information about slots in KB schema; on the other hand, it is flexible, as it does not use information about the values contained in the KB records.
This makes it easily adaptable to new values. 
These desirable properties are achieved by a separate decoder for each informable slot and a multi-label classifier for the requestable slots.
Those components explicitly assign values to slots like the fully structured approach, while also preserving the capability of dealing with out-of-vocabulary words like the free-form approach.
By using these two types of decoders, FSDM produces only valid belief states, overcoming the limitations of the free-form approach.
Further, FSDM has a new module called response slot binary classifier that adds extra supervision to generate the slots that will be present in the response more precisely before generating the final textual agent response
(see Section~\ref{methodology} for details).



The main contributions of this work are
\begin{enumerate}
\item FSDM, a task-oriented dialogue system with a new belief state tracking architecture that overcomes the limits of existing approaches and scales to real-world settings; 
\item  a new module, namely the response slot binary classifier, that helps to improve the performance of agent response generation;
\item FSDM achieves state-of-the-art results on both the Cambridge Restaurant dataset~\cite{wen2016network} and the Stanford in-car assistant dataset~\cite{eric2017key} without the need for fine-tuning through reinforcement learning
\end{enumerate}

\section{Related Work}

Our work is related to end-to-end task-oriented dialogue systems in general~\cite[among others]{BingNAACL18,Jason17,Lowe18,msr_challenge,BingGoogle17,Pawel18,bordes2016learning,HoriWHWHRHKJZA16,wen2016network,serban2016building} and those that extend the Seq2Seq~\cite{sutskever2014sequence} architecture in particular~\citep{eric2017key,madotto2018mem2seq, wen2018sequence}.
Belief tracking, which is necessary to form KB queries, is not explicitly performed in the latter works.
To compensate, 
\citet{eric2017key, XuH18, wen2018sequence} adopt a copy mechanism that allows copying information retrieved from the KB to the generated response.
\citet{madotto2018mem2seq} adopt Memory Networks~\citep{sukhbaatar2015end} to memorize the retrieved KB entities and words appearing in the dialogue history.
These models scale linearly with the size of the KB and need to be retrained at each update of the KB. 
Both issues make these approaches less practical in real-world applications.

\begin{figure*}[t]
\centering    
\includegraphics[width=0.85\linewidth]{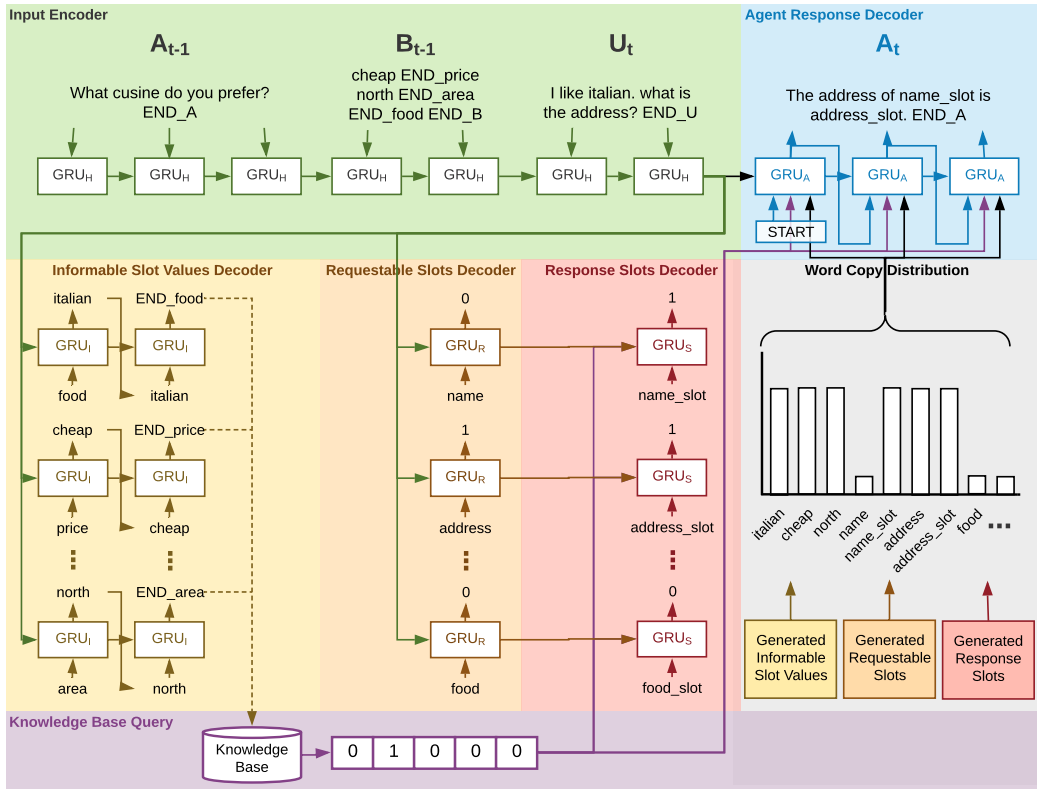}
\caption{FSDM architecture illustrated by a dialogue turn from the Cambridge Restaurant dataset with the following components: an input encoder (green), a belief state tracker (yellow for the informable slot values, orange for the requestable slots), a KB query component (purple), a response slot classifier (red), a component that calculates word copy probability (grey) and a response decoder (blue). Attention connections are not drawn for brevity.}
\label{fig:FSDM}
\end{figure*}

Our work is also akin to modularly connected end-to-end trainable networks~\citep{wen2016network, wen2017latent, BingNAACL18, BingGoogle17, msr_challenge, Zhong2018}.
\citet{wen2016network} includes belief state tracking and has two phases in training: the first phase uses belief state supervision, and then the second phase uses response generation supervision.
\citet{wen2017latent} improves \citet{wen2016network} by adding a policy network using latent representations so that the dialogue system can be continuously improved through reinforcement learning. 
These methods utilize classification as a way to decode the belief state.

\citet{leisequicity} decode the belief state as well as the response in a free-form fashion,
but it tracks the informable slot values without an explicit assignment to an informable slot.
Moreover, the arbitrary order in which informable slot values and requestable slots are encoded and decoded suggests that the sequential inductive bias the architecture provides may not be the right one.


Other works~\cite{Jang2016NeuralDS,Henderson2014RobustDS,Bapna2017TowardsZF,Kobayashi2018OutofDomainSV,XuACL2018} focus on the scalability of DST to large or changing vocabularies.
\citet{Rastogi2017} score a dynamically defined set of candidates as informable slot values.
\citet{DernoncourtLBB16} addresses the problem of large vocabularies with a mix of rules and machine-learned classifiers.


\section{Methodology}\label{methodology}

%
We propose a fully-fledged task-oriented dialogue system called \underline{F}lexibly-\underline{S}tructured \underline{D}ialogue \underline{M}odel (FSDM), which operates at the turn level.
Its overall architecture is shown in Figure~\ref{fig:FSDM}, which illustrates one dialogue turn. 
Without loss of generality, let us assume that we are on the $t$-th turn of a dialogue.
FSDM has three (3) inputs: agent response and belief state of the $t-1$-th turn, and user utterance of the $t$-th turn. 
It has two (2) outputs: the belief state for the $t$-th turn that is used to query the KB, and the agent response of the $t$-th turn based on the query result.
As we can see, belief tracking is the key component that turns unstructured user utterance and the dialogue history into a KB-friendly belief state.
The success of retrieving the correct KB result and further generating the correct response to complete a task relies on the quality of the produced belief state.






FSDM contains five (5) components that work together in an end-to-end manner as follows:
(1) The input is encoded and the last hidden state of the encoder serves as the initial hidden state of the belief state tracker and the response decoder;
(2) Then, the belief state tracker generates a belief state $B_t = \{I_t, R_t\}$, where $I_{t}$ is the set of constraints used for the KB query generated by the informable slots value decoder and $R_{t}$ is the user requested slots identified by the requestable slots multi-label classifier;
(3) Given $I_t$, the KB query component queries the KB and encodes the number of records returned in a one-hot vector $d_t$;
(4) The response slot binary classifier predicts which slots should appear in the agent response $S_t$;
(5) Finally, the agent response decoder takes in the KB output $d_t$, a word copy probability vector $\mathcal{P}^{c}$ computed from $I_t$, $R_t$, $S_t$ together with an attention on hidden states of the input encoder and the belief decoders, and generates a response $A_t$.

\subsection{Input Encoder}
The input contains three parts: (1) the agent response $A_{t-1}$, (2) the belief state $B_{t-1}$ from the $(t-1)$-th turn and (3) the current user utterance $U_t$.
These parts are all text-based and
concatenated, and then consumed by the input encoder.
Specifically, the belief state $B_{t-1}$ is represented as a sequence of informable slot names with their respective values and requestable slot names.
As an example, 
the sequence $\langle$\textit{cheap, end\_{price}, italian, end\_{food}, address, phone, end\_{belief}}$\rangle$ indicates a state where the user informed cheap and Italian as KB query constraints and requested the address and phone number.

The input encoder consists of an embedding layer followed by a recurrent layer with Gated Recurrent Units (GRU)~\cite{cho2014gru}.
It maps the input $A_{t-1} \circ B_{t-1} \circ U_{t}$ (where $\circ$ denotes concatenation) to a sequence of hidden vectors $\{h^{E}_i| i = 1, \dots, |A_{t-1} \circ B_{t-1} \circ U_{t}| \}$ so that $h^{E}_i = \text{GRU}_H(e^{A_{t-1} \circ B_{t-1} \circ U_{t}})$ where $e$ is the embedding function that maps from words to vectors.
The output of the input encoder is its last hidden state $h^{E}_{l}$, which is served as the initial state for the belief state and response decoders as discussed next.

\subsection{Informable Slot Value Decoder}
The belief state is composed of informable slot values $I_{t}$ and the requestable slots $R_{t}$.
We describe the generation of the former in this subsection and the latter in the next subsection.

The informable slot values track information provided by the user and are used to query the KB.
We allow each informable slot to have its own decoder to resolve the unwanted artificial dependencies among slot values introduced by TSCP \cite{leisequicity}. 
As an example of artificial dependency, `italian; expensive' appears a lot in the training data.
During testing, even when the gold informable value is `italian; moderate', the decoder may still generate `italian; expensive'. 
Modeling one decoder for each slot exactly associates the values with the corresponding informable slot.

The informable slot value decoder consists of GRU recurrent layers with a copy mechanism as shown in the yellow section of Figure~\ref{fig:FSDM}.
It is composed of weight-tied GRU generators that take the same initial hidden state $h^{E}_{l}$, but have different start-of-sentence symbols for each unique informable slot.
This way, each informable slot value decoder is dependent on the encoder's output, but it is also independent of the values generated for the other slots.
Let $\{k^{I}\}$ denote the set of informable slots. 
The probability of the $j$\textsuperscript{th} word $P(y^{k^I}_j)$ being generated for the slot $k^I$ is calculated as follows: (1) calculate the attention with respect to the input encoded vectors to obtain the context vector $c^{k^I}_j$, (2) calculate the generation score $\phi_g(y^{k^I}_j)$ and the copy score $\phi_c(y^{k^I}_j)$ based on the current step's hidden state $h^{k^I}_j$, (3) calculate the probability using the copy mechanism:
\begin{equation}
\small
    \begin{split}
        &c^{k^I}_j = \text{Attn}(h^{k^I}_{j-1}, \{h_{i}^E\}),\\
        &h^{k^I}_j = \text{GRU}_I\Big((c^{k^I}_j  \circ e^{y^{k^I}_{j}}), h^{k^I}_{j-1}\Big),\\
        &\phi_g(y^{k^I}_j) = W_{g}^{K^I}\cdot h^{k^I}_j,\\
        &\phi_c(y^{k^I}_j) = \text{tanh}(W_c^{K^I} \cdot h^{y_j^{k^I}}) \cdot h_j^{k^I} ,\\
        & y_j^{k^I} \in A_{t-1} \cup B_{t-1} \cup U_t,\\
        &P(y^{k^I}_j|y^{k^I}_{j-1}, h^{k^I}_{j-1}) = \text{Copy} \Big( \phi_c(y^{k^I}_j), \phi_g(y^{k^I}_j)\Big),
    \end{split}
\end{equation}
where for each informable slot $k^I$, $y_0^{k^I} = k^I$ and $h_0^{k^I} = h^{E}_{l}$, $e^{y^{k^I}_{j}}$ is the embedding of the current input word (the one generated at the previous step), and $W_{g}^{K^I}$ and $W_{c}^{K^I}$ are learned weight matrices.
We follow ~\cite{gu2016incorporating} and ~\cite{bahdanau2014neural} for the copy $\text{Copy}(\cdot, \cdot)$ and attention $\text{Attn}(\cdot, \cdot)$ mechanisms implementation respectively.

The loss for the informable slot values decoder is calculated as follows:
\begin{equation}
\small
\begin{split}
    \mathcal{L}^I =& - \frac{1}{|\{k^I\}|} \frac{1}{|Y^{k^I}|} \sum_{k^I} \sum_j \\
    &\log P(y^{k^I}_j = z^{k^I}_j|y^{k^I}_{j-1}, h^{k^I}_{j-1}),
\end{split}
\end{equation}
where $Y^{K^I}$ is the sequence of informable slot value decoder predictions and $z$ is the ground truth label.

\subsection{Requestable Slot Binary Classifier}
As the other part of a belief state, requestable slots are the attributes of KB entries that are explicitly requested by the user.
We introduce a separate multi-label requestable slots classifier to perform binary classification for each slot.
This greatly resolves the issues of TSCP that uses a single decoder with each step having unconstrained vocabulary-size choices, which may potentially lead to generating non-slot words.
Similar to the informable slots decoders, such a separate classifier also eliminates the undesired dependencies among slots. 


Let $\{k^R\}$ denote the set of requestable slots.
A single GRU cell is used to perform the classification.
The initial state $h^{E}_{l}$ is used to pay attention to the input encoder hidden vectors to compute a context vector $c^{k^R}$.
The concatenation of $c^{k^R}$ and $e^{k^R}$, the embedding vector of one requestable slot $k^R$, is passed as input and $h^{E}_{l}$ as the initial state to the GRU.
Finally, a sigmoid non-linearity is applied to the product of a weight vector $W_{y}^{R}$ and the output of the GRU $h^{k^R}$ to obtain $y^{k^R}$, which is the probability of the slot being requested by the user.
\begin{equation}
\small
\begin{split}
    &c^{k^R} = \text{Attn}(h^{E}_{l}, \{h_{i}^E\}),\\
    &h^{k^R} = \text{GRU}_R\Big( (c^{k^R}\circ e^{k^R}), h^{E}_{l} \Big),\\
    &y^{k^R} = \sigma (W_{y}^{R} \cdot h^{k^R}).
\end{split}
\end{equation}
The loss function for all requestable slot binary classifiers is:
\begin{equation}
\small
    \begin{split}
    \mathcal{L}^R =& - \frac{1}{|\{k^R\}|} \sum_{k^R} \\
    &z^{k^R} \log (y^{k^R}) + (1-z^{k^R}) \log (1-y^{k^R}).
    \end{split}
\end{equation}

\subsection{Knowledge Base Query}
The generated informable slot values $I_t = \{Y^{k^I}\}$ are used as constraints of the KB query.
The KB is composed of one or more relational tables and each entity is a record in one table.        
The query is performed to select a subset of the entities that satisfy those constraints.
For instance, if the informable slots are \{\textit{price=cheap, area=north}\}, all the restaurants that have attributes of those fields equal to those values will be returned.
The output of this component, the one-hot vector $d_t$, indicates the number of records satisfying the constraints.
$d_t$ is a five-dimensional one-hot vector, where the first four dimensions represent integers from 0 to 3 and the last dimension represents 4 or more matched records.
It is later used to inform the response slot binary classifier and the agent response decoder.

\subsection{Response Slot Binary Classifier}

In order to incorporate all the relevant information about the retrieved entities into the response, FSDM introduces a new response slot binary classifier.
Its inputs are requestable slots and KB queried result $d_t$
and the outputs are the response slots to appear in the agent response.
Response slots are the slot names that are expected to appear in a de-lexicalized response (discussed in the next subsection).
For instance, 
assume the requestable slot in the belief state is ``address'' and the KB query returned one candidate record.
The response slot binary classifier may predict {name\_slot}, {address\_slot} and {area\_slot}, which are expected to appear in an agent response as ``{name\_slot} is located in {address\_slot} in the {area\_slot} part of town''\footnote{
Before the agent response is presented to the user, those slot names are replaced by the actual values of the KB entries.}.


The response slots $\{k^S\}$ map one-to-one to the requestable slots $\{k^R\}$.
The initial state of each response slot decoder is the last hidden state of the corresponding requestable slot decoder.
In this case, the context vector $c^{k^S}$ is obtained by paying attention to all hidden vectors in the informable slot value decoders and requestable slots classifiers.
Then, the concatenation of the context vector $c^{k^S}$, the embedding vector of the response slot $e^{k^S}$ and the KB query vector $d_t$ are used as input to a single GRU cell.
Finally, a sigmoid non-linearity is applied to the product of a weight vector $W_{y}^{S}$ and the output of the GRU $h^{k^S}$ to obtain a probability $y^{k^S}$ for each slot that is going to appear in the answer.
\begin{equation}
\small
\begin{split}
    &c^{k^S} = \text{Attn}(h^{k^R}, \\
    &\{h_{i}^{k^I}|k^I \in K^I, i \le |Y^{k^I}|\} \cup \{h^{k^R}| k^R \in K^R\}), \\
    &h^{k^S} = \text{GRU}_S\Big((c^{k^S} \circ e^{k^S} \circ d_t), h^{k^R}\Big),\\
    &y^{k^S} = \sigma (W_{y}^{S} \cdot h^{k^S}).
\end{split}
\end{equation}
The loss function for all response slot binary classifiers is:
\begin{equation}
\small
\begin{split}
    \mathcal{L}^S =& - \frac{1}{|\{k^S\}|} \sum_{k^S} \\
    &z^{k^S} \log (y^{k^S}) + (1-z^{k^S}) \log (1-y^{k^S}).
\end{split}
\end{equation}

\subsection{Word Copy Probability and Agent Response Decoder}
Lastly, we introduce the agent response decoder. 
It takes in the generated informable slot values, requestable slots, response slots, and KB query result and generates a (de-lexicalized) response.
We adopt a copy-augmented decoder~\cite{gu2016incorporating} as architecture.
The canonical copy mechanism only takes a sequence of word indexes as inputs but does not accept the multiple Bernoulli distributions we obtain from sigmoid functions.
For this reason, we introduce a vector of independent word copy probabilities $\mathcal{P}^{C}$, which is constructed as follows:

\begin{equation}
\small
\mathcal{P^C}(w) = \begin{cases}
    y^{k^R}, & \text{if } w = k^R,\\
    y^{k^S}, & \text{if } w = k^S,\\
    1, & \text{if } w \in I_t,\\
    0, & \text{otherwise},
  \end{cases}
\end{equation}
where if a word $w$ is a requestable slot or a response slot, the probability is equal to their binary classifier output; if a word appears in the generated informable slot values, its probability is equal to $1$; for the other words in the vocabulary the probability is equal to $0$.
This vector is used in conjunction with the agent response decoder prediction probability to generate the response.


The agent response decoder is responsible for generating a de-lexicalized agent response.
The response slots are substituted with the values of the results obtained by querying the KB before the response is returned to the user.

Like the informable slot value decoder, the agent response decoder also uses a copy mechanism, so it has a copy probability and generation probability.
Consider the generation of the $j$\textsuperscript{th} word.
Its generation score $\phi_g$ is calculated as:
\begin{equation}
\small
    \begin{split}
        &c^{A^E}_j = \text{Attn}(h_{j-1}^A, \{h_i^E\}), \\
        &c^{A^B}_j = \text{Attn}(h_{j-1}^A, \{h_{i}^{k^I}|k^I \in K^I, i \le |Y^{k^I}|\}\\ 
        &\cup \{h^{k^R}| k^R \in K^R\})\cup \{h^{k^S}| k^S \in K^S\}),\\
        &h^{A}_j = \text{GRU}_A\Big( (c^{A^E}_j \circ  c^{A^B}_j \circ e^{A}_j \circ d_t), h_{j-1}^A \Big),\\
        &\phi_g(y^A_j) = W_{g}^{A} \cdot h^{A}_j,
    \end{split}
\end{equation}
where $c^{A^E}_j$ is a context vector obtained by attending to the hidden vectors of the input encoder, $c^{A^B}_j$ is a context vector obtained by attending to all hidden vectors of the informable slot value decoder, requestable slot classifier and response slot classifier, and $W_{g}^{A}$ is a learned weight matrix.
The concatenation of the two context vectors $c^{A^E}_j$ and $c^{A^B}_j$, the embedding vector $e^{A}_j$ of the previously generated word and the KB query output vector $d_t$ is used as input to a GRU.
Note that the initial hidden state is $h_0^A = h^{E}_{l}$.
The copy score $\phi_c$ is calculated as:
\begin{equation}
\small
    \phi_c(y_j^A) = \begin{cases}
    \mathcal{P}^C(y_j^A) \cdot \text{tanh}(W_c^A \cdot h^{y_j^A}) \cdot h_j^A, &\\
    \text{if } y_j^A \in I_t \cup K^R \cup K^S,&\\
    \mathcal{P}^C(y_j^A),  \text{otherwise},&
    \end{cases}
\end{equation}
where $W_c^A$ is a learned weight matrix.
The final probability is:
\begin{equation}
\small
    P(y^{A}_j|y^{A}_{j-1}, h^{A}_{j-1}) = \text{Copy}(\phi_g(y^A_j), \phi_c(y_j^A)).
\end{equation}
Let $z$ denote the ground truth de-lexicalized agent response.
The loss for the agent response decoder is calculated as follows where $Y^A$ is the sequence of agent response decoder prediction:
\begin{equation}
\small
    \mathcal{L}^A = - \frac{1}{|Y^{A}|}  \sum_j \log P(y^{A}_j = z^{A}_j|y^{A}_{j-1}, h^{A}_{j-1}).
\end{equation}

\subsection{Loss Function}
The loss function of the whole network is the sum of the four losses described so far for the informable slot values $\mathcal{L}^I$, requestable slot $\mathcal{L}^R$, response slot $\mathcal{L}^S$ and agent response decoders $\mathcal{L}^A$, weighted by $\alpha$ hyperparameters:
\begin{equation}
\small
    \mathcal{L} = \alpha^{I}\mathcal{L}^I + \alpha^{R}\mathcal{L}^R + \alpha^{S}\mathcal{L}^S +
    \alpha^{A}\mathcal{L}^A.
    \label{eq:loss}
\end{equation}
The loss is optimized in an end-to-end fashion, with all modules trained simultaneously with loss gradients back-propagated to their weights.
In order to do so, ground truth results from database queries are also provided to the model to compute the $d_t$, while at prediction time results obtained by using the generated informable slot values $I_t$ are used.

\section{Experiments}
\label{sec:experiment}
\begin{table}
  \centering
  \resizebox{\columnwidth}{!}{
  \begin{tabular}{c|l|l|l}
  \multicolumn{4}{c}{CamRest: restaurant reservation}\\\hline
  dialogue split&train: 408 &dev: 136    &test: 136\\\hline
  \# of keys&informable: 3& requestable: 7 &response: 7\\\hline
  database record& \multicolumn{3}{l}{99} \\\hline
  \multicolumn{4}{c}{KVRET: navigation, weather, calendar scheduling}\\\hline
  dialogue split&train: 2425 &dev: 302    &test: 302\\\hline
  \# of keys&informable: 10& requestable: 12 &response: 12\\\hline
  database record& \multicolumn{3}{l}{284} \\\hline
  \end{tabular}
  }
  \caption{Dataset}
  \label{tab:dataset}
\end{table}
\begin{table*}
  \centering
  \resizebox{\linewidth}{!}{
  \begin{tabular}{l|lll|lll|lll|lll}
  Dataset  & \multicolumn{6}{|c|}{CamRest} & \multicolumn{6}{|c}{KVRET} \\\hline
  Method  & Inf P & Inf R & Inf F$_1$ & Req P & Req R & Req F$_1$ & Inf P & Inf R & Inf F$_1$ & Req P & Req R & Req F$_1$ \\\hline
    TSCP/RL$^\dagger$ &0.970    &0.971    &0.971    &0.983    &0.935    &0.959 &\textbf{0.936}    &0.874    &0.904    &0.725    &0.485    &0.581\\
    TSCP$^\dagger$  &0.970    &0.971    &0.971    &0.983    &0.938    &0.960 &0.934    &0.890    &0.912    &0.701    &0.435    &0.526\\
  \hline
  FSDM/Res & 0.979 & 0.984 & 0.978 & 0.994 & 0.947 & 0.967 & 0.918 & 0.930 & 0.925 & 0.812 & 0.993 & 0.893\\
  FSDM & \textbf{0.983}* & \textbf{0.986}* & \textbf{0.984}* & \textbf{0.997}* & \textbf{0.952} & \textbf{0.974}* & 0.92 & \textbf{0.935}* & \textbf{0.927}* & \textbf{0.819}* & \textbf{1.000}* & \textbf{0.900}* \\
  \hline
  \end{tabular}
  }
  \caption{Turn-level performance results.
  \textbf{Inf}: Informable, \textbf{Req}: Requestable, \textbf{P}: Precision, \textbf{R}: Recall.
  Results marked with $\dagger$ are computed using available code, and all the other ones are reported from the original papers. $*$ indicates the improvement is statistically significant with $p=0.05$.
  }
  \label{tab:turnresult}
\end{table*}
\begin{table}
  \centering
  \resizebox{\columnwidth}{!}{
  \begin{tabular}{l|lll|lll}
  Dataset  & \multicolumn{3}{|c|}{CamRest} & \multicolumn{3}{|c}{KVRET} \\\hline
  Method  & BLEU & EMR & SuccF$_1$  & BLEU & EMR & SuccF$_1$\\\hline
  NDM     & 0.212 & 0.904 & 0.832 &0.186 &0.724 &0.741\\
  LIDM    & 0.246 & 0.912 & 0.840 &0.173 &0.721 &0.762\\
  KVRN    & 0.134 & -     & -     &0.184 &0.459 &0.540\\
  TSCP    & 0.253 & 0.927 & 0.854 &\textbf{0.219} &0.845 &0.811\\\hline
  TSCP/RL $^\dagger$ &0.237    &0.915    &0.826 &0.195    &0.809    &0.814\\
  TSCP$^\dagger$ &0.237    &0.913    &0.841 &0.189 &0.833    &0.81\\
  \hline
  FSDM/St &0.245 &- &0.847 &0.204 &- &0.809\\
  FSDM/Res & 0.251 & 0.924 & 0.855 & 0.209 & 0.834 & 0.815\\
  FSDM & \textbf{0.258}* & \textbf{0.935}* & \textbf{0.862}* & 0.215 & \textbf{0.848}* & \textbf{0.821}* \\\hline
  
  \end{tabular}
  }
  \caption{Dialogue level performance results.
   \textbf{SuccF$_1$}: Success F$_1$ score, \textbf{EMR}: Entity Match Rate.
  Results marked with $\dagger$ are computed using available code, and all the other ones are reported from the original papers. $*$ indicates the improvement is statistically significant with $p=0.05$.
  }
  \label{tab:diaresult}
\end{table}
\begin{table}[t]
\centering
\resizebox{\columnwidth}{!}{
  \begin{tabular}{ll}
 
user msg& what is the date and time of \\
&my next meeting and who will be attending it ?\\
\hline
&\textbf{belief state}\\
\hline
\textbf{GOLD}& informable slot (event=meeting),\\
&requestable slot (date, time, party)\\
TSCP& `meeting' `$\langle$EOS\_Z1$\rangle$' `date' `;' `party'\\
FSDM& event=meeting date=True time=True party = True\\
\hline
& \textbf{agent response}\\
\hline
\textbf{GOLD}& your next meeting is with \\
&party\_SLOT on the date\_SLOT at time\_SLOT. \\
TSCP& your next meeting is at time\_SLOT \\
&on date\_SLOT at time\_SLOT . \\
FSDM& you have a meeting on date\_SLOT \\
&at time\_SLOT with party\_SLOT\\
\hline

\end{tabular}
}
\caption{Example of generated belief state and response for calendar scheduling domain}
\label{tab:dialogue}
\end{table}

We tested the FSDM on the Cambridge Restaurant dataset (CamRest)~\cite{wen2016network} and the Stanford in-car assistant dataset (KVRET)~\cite{eric2017key} described in Table~\ref{tab:dataset}.
\subsection{Preprocessing and Hyper-parameters}


We use NLTK~\cite{Bird2009NLP} to tokenize each sentence.
The user utterances are precisely the original texts, while all agent responses are de-lexicalized as described in \cite{leisequicity}.
We obtain the labels for the response slot decoder from the de-lexicalized response texts.
We use 300-dimensional GloVe embeddings~\cite{pennington2014glove} trained on 840B words.
Tokens not present in GloVe are initialized to be the average of all other embeddings plus a small amount of random noise to make them different from each other. 

We optimize both training and model hyperparameters by running Bayesian optimization over the product of validation set BLEU, EMR, and SuccF$_1$ using skopt\footnote{\url{https://scikit-optimize.github.io/}}.
The model that performed the best on the validation set uses Adam optimizer~\cite{kingma2014adam} with a learning rate of 0.00025 for minimizing the loss in Equation~\ref{eq:loss} for both datasets.
We apply dropout with a rate of 0.5 after the embedding layer, the GRU layer and any linear layer for CamRest and 0.2 for KVRET.
The dimension of all hidden states is 128 for CamRest and 256 for KVRET.
Loss weights $\alpha^I$, $\alpha^R$, $\alpha^S$, $\alpha^A$ are 1.5, 9, 8, 0.5 respectively for CamRest and 1, 3, 2, 0.5 for KVRET.

\subsection{Evaluation Metrics}
We evaluate the performance concerning belief state tracking, response language quality, and task completion.
For belief state tracking, we report precision, recall, and F$_1$ score of informable slot values and requestable slots.
BLEU~\cite{papineni2002bleu} is applied to the generated agent responses for evaluating language quality.
Although it is a poor choice for evaluating dialogue systems~\cite{LiuLSNCP16}, we still report it in order to compare with previous work that has adopted it.
For task completion evaluation, the Entity Match Rate (EMR)~\cite{wen2016network} and Success F$_1$ score (SuccF$_1$)~\cite{leisequicity} are reported.
EMR evaluates whether a system can correctly retrieve the user's indicated entity (record) from the KB based on the generated constraints so it can have only a score of 0 or 1 for each dialogue.
The SuccF$_1$ score evaluates how a system responds to the user's requests at dialogue level: it is F$_1$ score of the response slots in the agent responses.

\subsection{Benchmarks}

We compare FSDM with four baseline methods and two ablations.

\textbf{NDM}~\cite{wen2016network} proposes a modular end-to-end trainable network.
It applies de-lexicalization on user utterances and responses.

\textbf{LIDM}~\cite{wen2017latent} improves over NDM by employing a discrete latent variable to learn underlying dialogue acts.
This allows the system to be refined by reinforcement learning.

\textbf{KVRN}~\cite{eric2017key} adopts a copy-augmented Seq2Seq model for agent response generation and uses an attention mechanism on the KB.
It does not perform belief state tracking.

\textbf{TSCP/RL}~\cite{leisequicity} is a two-stage CopyNet which consists of one encoder and two copy-mechanism-augmented decoders for belief state and response generation.
\textbf{TSCP} includes further parameter tuning with reinforcement learning to increase the appearance of response slots in the generated response.
We were unable to replicate the reported results using the provided code\footnote{\url{https://github.com/WING-NUS/sequicity}}, hyperparameters, and random seed,
so we report both the results from the paper and the average of 5 runs on the code with different random seeds (marked with $^\dagger$).


\textbf{FSDM} is the proposed method and we report two ablations: in \textbf{FSDM/St} the whole state tracking is removed (informable, requestable and response slots) and the answer is generated from the encoding of the input, while in \textbf{FSDM/Res}, only the response slot decoder is removed.

\subsection{Result Analysis}
At the turn level, FSDM and FSDM/Res perform better than TSCP and TSCP/RL on belief state tracking, especially on requestable slots, as shown in Table~\ref{tab:turnresult}.
FSDM and FSDM/Res use independent binary classifiers for the requestable slots and are capable of predicting the correct slots in all those cases.
FSDM/Res and TSCP/RL do not have any additional mechanism for generating response slot, so FSDM/Res performing better than TSCP/RL shows the effectiveness of flexible-structured belief state tracker.
Moreover, FSDM performs better than FSDM/Res, but TSCP performs worse than TSCP/RL.
This suggests that using RL to increase the appearance of response slots in the response decoder does not help belief state tracking, but our response slot decoder does. 

FSDM performs better than all benchmarks on the dialogue level measures too, as shown in Table~\ref{tab:diaresult}, with the exception of BLEU score on KVRET, where it is still competitive.
Comparing TSCP/RL and FSDM/Res, the flexibly-structured belief state tracker achieves better task completion than the free-form belief state tracker.
Furthermore, FSDM performing better than FSDM/Res shows the effectiveness of the response slot decoder for task completion.
The most significant performance improvement is obtained on CamRest by FSDM, confirming that the additional inductive bias helps to generalize from smaller datasets.
More importantly, the experiment confirms that, although making weaker assumptions that are reasonable for real-world applications, FSDM is capable of performing at least as well as models that make stronger limiting assumptions which make them unusable in real-world applications.



\subsection{Error Analysis}

We investigated the errors that both TSCP and FSDM make and discovered that the sequential nature of the TSCP state tracker leads to the memorization of common patterns that FSDM is not subject to.
As an example (Table~\ref{tab:dialogue}), TSCP often generates ``date; party'' as requestable slots even if only ``party'' and ``time'' are requested like in ``what time is my next activity and who will be attending?'' or if ``party'', ``time'' and ``date'' are requested like in ``what is the date and time of my next meeting and who will be attending it?''.
FSDM produces correct belief states in these examples.

FSDM misses some requestable slots in some conditions.
For example, consider the user's utterance: ``I would like their address and what part of town they are located in''.
The ground-truth requestable slots are `address' and `area'.
FSDM only predicts `address' and misses `area', which suggests that the model did not recognize `what part of town' as being a phrasing for requesting  `area'. 
Another example is when the agent proposes ``the name\_SLOT is moderately priced and in the area\_SLOT part of town . would you like their location ?'' and the user replies ``i would like the location and the phone number, please''.
FSDM predicts `phone' as a requestable slot, but misses `address', suggesting it doesn't recognize the connection between `location' and `address'.
The missing requestable slot issue may propagate to the agent response decoder. 
These issues may arise due to the use of fixed pre-trained embeddings and the single encoder.
Using separate encoders for user utterance, agent response and dialogue history or fine-tuning the embeddings may solve the issue.

\section{Conclusion}
We propose the flexibly-structured dialogue model, a novel end-to-end architecture for task-oriented dialogue.
It uses the structure in the schema of the KB to make architectural choices that introduce inductive bias and address the limitations of fully structured and free-form methods.
The experiment suggests that this architecture is competitive with state-of-the-art models, while at the same time providing a more practical solution for real-world applications.


\section*{Acknowledgments}
We would like to thank Alexandros Papangelis, Janice Lam, Stefan Douglas Webb and SIGDIAL reviewers for their valuable comments.

\bibliography{dialogue2018}
\bibliographystyle{emnlp_natbib}

\end{document}